\documentclass[letterpaper]{article} 
\usepackage{aaai2027}
\nocopyright

\usepackage[hyphens]{url} 
\usepackage{graphicx}    
\usepackage{natbib}      
\usepackage{caption}     
\usepackage{amsmath}
\usepackage{amssymb}
\usepackage{siunitx}
\usepackage{pifont}
\usepackage{adjustbox}
\usepackage{booktabs}
\usepackage{multirow}
\usepackage{tabularx}
\usepackage{array}

\usepackage{subcaption}

\usepackage[table]{xcolor}

\usepackage[strict]{changepage}
\usepackage{framed}

\definecolor{formalshade}{rgb}{0.85,1,0.85}
\definecolor{darkgreen}{rgb}{0.0,0.6,0.30}

\newenvironment{answer}{%
  \MakeFramed{\advance\hsize-\width\FrameRestore}%
  \noindent
  \hspace{-4.55pt}%
  \begin{adjustwidth}{}{7pt}%
}{%
  \end{adjustwidth}%
  \endMakeFramed%
}

\usepackage{xspace}
\usepackage{acronym}

\newcommand{\wcircle}[1]{\ding{\numexpr171 + #1}}
\newcommand{\bcircle}[1]{\ding{\numexpr181 + #1}}

\title{Old Tricks, New Models: How Simple Image Transformations \\ Break Modern AI-based Content Moderation}

\author{
Marco Alecci\textsuperscript{\rm 1},
Francesco Marchiori\textsuperscript{\rm 2},
Iyiola Emmanuel Olatunji\textsuperscript{\rm 1},
Tegawendé F. Bissyandé\textsuperscript{\rm 1},
and Jacques Klein\textsuperscript{\rm 1}
}
\affiliations{
\textsuperscript{\rm 1}Interdisciplinary Centre for Security, Reliability and Trust (SnT), University of Luxembourg, Luxembourg\\
\textsuperscript{\rm 2}Department of Mathematics, University of Padua, Padua, Italy\\
marco.alecci@uni.lu, francesco.marchiori.4@phd.unipd.it, emmanuel.olatunji@uni.lu,\\
tegawende.bissyande@uni.lu, jacques.klein@uni.lu
}

\begin{document}
\maketitle

\begin{abstract}
While automated content-moderation systems have become essential for screening harmful content at scale, conventional task-specific classifiers often provide limited policy coverage and contextual understanding. Recently, commercial multimodal moderation APIs built on large foundation models have been introduced with the promise of providing broader and more capable safety filters. 
In this work, we analyze whether this shift also yields more robust image moderation. We conduct a large-scale black-box evaluation on three established commercial image-moderation services and compare their robustness.
By evaluating seven simple, model-agnostic image transformations across multiple providers, datasets, harm categories, perceptual-similarity constraints, and transformation intensities, we find that: (1) all three commercial services can be bypassed using inexpensive image transformations that require no gradients, surrogate models, or knowledge of the target system; (2) even fixed transformations such as color inversion and grayscale conversion induce unsafe-to-safe decision changes while preserving content that remains recognizable to humans; (3) their robustness varies substantially across datasets and harm categories, with multimodal content and self-harm exhibiting pronounced vulnerabilities. This yields the conclusion that replacing conventional moderation classifiers with foundation-model-based APIs does not, by itself, provide a reliable security boundary. Such systems must be evaluated under realistic transformations and deployed as one component of a layered moderation pipeline rather than as standalone safety filters. 
\\ 

\textit{\textbf{\color{red}Warning:} This paper contains sensitive images of unsafe content that readers may find disturbing.}
\end{abstract}

\section{Introduction}
\label{sec:introduction}

Automated content moderation has become a critical layer of modern social media platforms, where large volumes of user-generated text, images, and videos are rapidly shared and algorithmically recommended. Without effective safeguards, sexually explicit, violent, hateful, or otherwise harmful material may reach broad audiences, including children and teenagers, before it can be reviewed or removed, as reported by numerous journalistic investigations~\cite{horwitz2024instagram,adams2021sexualimages,das2022instagram,lomas2023metachildprotection}. At this scale, purely manual moderation is insufficient because it is costly, slow, difficult to apply consistently, and exposes human moderators to harmful material~\cite{roberts2019behind,gillespie2018custodians}. Automated moderation is therefore essential for screening content at the speed and scale required by modern platforms.

To address these challenges, researchers and companies have developed automated moderation systems for text~\cite{perspectiveapi2026,malik2025deep,googlecloudenaturallanguageapi,caselli2021hatebert,sarkar2021fbert}, images~\cite{pandey2021device,googlecloud_vision_safesearch_2026,schuhmann_clip_nsfw_detector_2022,yahoo_open_nsfw}, and multimodal content~\cite{aws_rekognition_moderation,microsoft_azure_ai_content_safety_overview,kiela2020hateful,das2020detecting}. While many earlier approaches rely on task-specific classifiers, recent foundation model-based systems provide broader policy coverage and richer contextual understanding~\cite{mistral2026moderation,zeng2024shieldgemma,inan2023llamaguard,helff2024llavaguard,zeng2025shieldgemma,openai2024omnimoderation}. A prominent example is OpenAI's \texttt{omni-moderation-latest}, a text-and-image moderation API described as being built on GPT-4o~\cite{openai2024omnimoderation}. Since it is offered free of charge under current usage limits, it is particularly attractive and accessible to small developers and services without dedicated moderation infrastructure. Weaknesses in such a widely deployable API may therefore directly affect downstream applications that rely on it as a first line of defense.

Adversarial attacks and robustness evaluations have been extensively studied for traditional machine-learning models, including image classifiers~\cite{szegedy2013intriguing,goodfellow2014explaining,madry2017towards}, but much less is known about modern commercial moderation systems. Their black-box nature makes traditional attack construction difficult, since attackers cannot directly optimize against the target model and may not have access to a well-matched surrogate.
Prior work has shown that surrogate-based attacks can succeed~\cite{papernot2017practical}, but also that transferability degrades under realistic mismatches in dataset source, model architecture, and class balance~\cite{alecci2023your,marchiori2025dumb}. At the same time, effective evasion does not always require gradients or optimized adversarial examples. In particular, simple, model-agnostic image transformations can already act as practical offensive strategies~\cite{alecci2023your,marchiori2025dumb}. This motivates our central question: \emph{How stable are commercial image-moderation systems under simple, model-agnostic visual transformations?}


In this work, we investigate whether widely deployed commercial image-moderation services remain reliable under simple, model-agnostic visual transformations. We conduct a black-box robustness evaluation of OpenAI omni-moderation, Amazon Rekognition, and Google Cloud SafeSearch using seven transformations that require no gradients, surrogate models, or knowledge of the target systems. We first compare the three services under the same controlled setting using pornographic images. We then examine OpenAI omni-moderation at a larger scale across three datasets covering pornographic, violent, self-harm, and multimodal text-image content, studying robustness across harm categories, perceptual-similarity constraints, and transformation intensities. 
Overall, the evaluation comprises approximately \num{600000} moderation API queries. Our results show that all three commercial services are vulnerable to inexpensive one-shot and intensity-dependent transformations. Unsafe-to-safe decision changes occur even when the transformed image remains substantially similar to the original. The larger analysis also shows that robustness varies considerably across content types and harm categories, and that many decisions change at low or moderate transformation intensities. These findings indicate that the vulnerabilities are not limited to a single provider and that commercial moderation APIs should not be treated as standalone security boundaries.

\noindent
\textbf{Contributions.}
We make the following contributions:
\begin{itemize}
    \item  We conduct a large-scale black-box robustness evaluation of three commercial image-moderation APIs under seven model-agnostic transformations.
    \item  We show that inexpensive one-shot and intensity-dependent transformations can bypass all three services without gradients, surrogate models, or target-system knowledge, including under non-trivial image-similarity constraints.
    \item We characterize robustness across providers, unsafe-content types, similarity constraints, and transformation intensities, and release our experimental artifacts to support reproducibility and further research.
\end{itemize}

\section{Related Work}
\label{sec:related_work}
In this section, we review prior work related to our study.

\subsection{Automated Content Moderation}
Automated content moderation is widely used to screen unsafe user-generated content at scale. Existing systems include text classifiers for toxicity and hate speech~\cite{perspectiveapi2026,caselli2021hatebert,sarkar2021fbert}, image classifiers for nudity and unsafe visual content~\cite{googlecloud_vision_safesearch_2026,schuhmann_clip_nsfw_detector_2022,yahoo_open_nsfw}, and multimodal models for image--text moderation~\cite{kiela2020hateful,das2020detecting}. Commercial moderation APIs are particularly attractive because they provide scalable policy enforcement and simple integration without requiring organizations to train and maintain dedicated models. Recent moderation systems increasingly incorporate foundation models and multimodal architectures. Examples include LLM-based guardrails such as Llama Guard, LLaVAGuard, and ShieldGemma~\cite{inan2023llamaguard, helff2024llavaguard, zeng2024shieldgemma} or OpenAI's multimodal \texttt{omni-moderation-latest} API~\cite{openai2024omnimoderation}. A related research direction studies safety in text-to-image generation, where unsafe prompts or generated outputs are filtered or modified~\cite{yuan2026promptguard}. In contrast, we study downstream image moderation, where an already-produced image is classified as safe or unsafe by a deployed moderation service.

\subsection{Adversarial Attacks on Image Classifiers}
Adversarial image classifiers can be fooled by small, carefully designed perturbations, including gradient-based attacks such as FGSM and stronger robust-optimization methods~\cite{szegedy2013intriguing,goodfellow2014explaining,madry2017towards}. These approaches often assume white-box access, whereas black-box attacks rely on substitute models and adversarial transferability~\cite{papernot2017practical}. Later work showed that transferability can substantially degrade under realistic mismatches in dataset source, model architecture, and class balance, while also demonstrating that simple, non-optimized image transformations can remain effective offensive strategies~\cite{alecci2023your,marchiori2025dumb}. Related robustness studies evaluate common corruptions and realistic transformations, including noise, blur, compression, and digital artifacts~\cite{hendrycks2019benchmarking,hendrycks2020augmix,tian2018deeptest}. Building on this practical perspective, we do not optimize perturbations against the target systems; instead, we evaluate whether deployed image-moderation services remain stable under simple, model-agnostic transformations.

\subsection{Adversarial Attacks and Audits of Content Moderation}
Prior evaluations of commercial moderation systems have focused mainly on text, examining performance, linguistic variation, bias, and over- or under-moderation across social groups~\cite{markov2023holistic,hartmann2025lost}. Text-based attacks have likewise shown that simple edits and perceptual mismatches, including typos, typographic manipulations, and visual smuggling, can evade deployed moderation systems~\cite{grondahl2018all,yang2026eyes,li2026making}. In contrast, the robustness of deployed commercial image-moderation services to simple visual transformations remains comparatively underexplored. Prior evaluations of these transformations have primarily considered general image-classification tasks rather than deployed moderation systems~\cite{alecci2023your,marchiori2025dumb}. We address this gap through a black-box evaluation of OpenAI \texttt{omni-moderation}, Amazon Rekognition, and Google Cloud SafeSearch.
\section{Experimental Setup}
\label{sec:experimental_setup}
In this section, we describe the experimental setup used in our evaluation. 

\subsection{Target Systems}
\label{sec:targetSystems}
We evaluate three widely available commercial image-moderation services: \wcircle{1} OpenAI \texttt{omni-moderation}~\cite{openai2024omnimoderation,openai_moderation_docs}, \wcircle{2} Amazon Rekognition~\cite{aws_rekognition_moderation}, and \wcircle{3} Google Cloud SafeSearch~\cite{googlecloud_vision_safesearch_2026}. All three provide black-box API access and return category-level moderation outputs, although they differ in taxonomy and output format. OpenAI describes \texttt{omni-moderation} as being built on GPT-4o~\cite{openai2024omnimoderation}, whereas Amazon and Google do not publicly disclose sufficient architectural details to characterize their underlying models/architecture. We therefore refer to the three systems collectively as commercial AI-based image-moderation services. 

\noindent\bcircle{1} \textbf{OpenAI \texttt{omni-moderation}.}
The API supports both text and image inputs and returns a global binary decision, \texttt{flagged}, together with category-specific boolean labels and confidence scores. For image inputs, supported categories include \texttt{sexual}, \texttt{violence}, \texttt{violence/graphic}, and \texttt{self-harm}; other categories, such as \texttt{hate} and \texttt{harassment}, are documented as text-only~\cite{openai_moderation_docs}.

\noindent\bcircle{2} \textbf{Amazon Rekognition.}
Amazon Rekognition analyzes JPEG and PNG images through the \texttt{DetectModerationLabels} operation. It uses a three-level hierarchical taxonomy and returns labels such as \texttt{Explicit Nudity}, \texttt{Sexual Activity}, \texttt{Graphic Violence}, and \texttt{Self-Harm}, together with confidence scores~\cite{aws_rekognition_moderation}. 

\noindent\bcircle{3} \textbf{Google Cloud SafeSearch.}
Google Cloud SafeSearch is an explicit-content detection feature of the Cloud Vision API. It evaluates five categories: \texttt{adult}, \texttt{spoof}, \texttt{medical}, \texttt{violence}, and \texttt{racy}; and assigns each one a likelihood ranging from \texttt{VERY\_UNLIKELY} to \texttt{VERY\_LIKELY}~\cite{googlecloud_vision_safesearch_2026}.

OpenAI currently provides access to \texttt{omni-moderation} free of charge under its usage policies~\cite{openai_moderation_docs}, whereas Amazon Rekognition and Google Cloud SafeSearch are paid services, although both may offer limited free-usage tiers. This difference also has practical implications: as a freely accessible multimodal moderation API, \texttt{omni-moderation} may be more readily adopted as a first line of defense for user-generated content, making its robustness particularly relevant to real-world deployments.

\subsection{Adversarial Modifications}
\label{sec:adversarialModifications}

To assess the robustness of the target moderation systems, we consider a suite of simple, model-agnostic image transformations inspired by prior work on so-called ``non-mathematical'' attacks~\cite{alecci2023your,marchiori2025dumb}. The transformations are divided into two groups: \wcircle{1} \textit{intensity-dependent transformations}, whose effect can be progressively adjusted through a tunable parameter; and \wcircle{2} \textit{one-shot transformations}, which apply a fixed modification to the input image. For intensity-dependent transformations, we denote the intensity parameter generically as $\epsilon$, although its meaning and numerical scale depend on the specific transformation. Using the PIL\footnote{\url{https://pypi.org/project/pillow/}} Python library, we implemented the following transformations:
\begin{itemize}
    \item \textit{Box Blur} -- We blur the image by replacing each pixel with the average value of the pixels in a square neighborhood around it. The intensity $\epsilon$ controls the blur radius.
    \item \textit{Gaussian Noise} -- We add random noise sampled from a Gaussian distribution to the image pixels. The intensity $\epsilon$ controls the standard deviation of the noise.
    \item \textit{Gaussian Blur} -- We blur the image by convolving it with a Gaussian kernel, which gives greater weight to pixels near the center of the neighborhood. The intensity $\epsilon$ controls the blur radius.
    \item \textit{Salt and Pepper} -- We randomly replace a fraction of image pixels with either black or white pixels, creating sparse impulse noise. The intensity $\epsilon$ controls the proportion of corrupted pixels.
    \item \textit{Split-Merge RGB} -- We split the image into its red, green, and blue color channels and merge them back in a different order, altering the image colors while preserving its structure. This transformation does not depend on $\epsilon$.
    \item \textit{Grayscale Filter} -- We remove color information from the image and retain only luminance values, producing an image composed of shades of gray. This transformation does not depend on $\epsilon$.
    \item \textit{Invert Color} -- We produce a negative version of the image by subtracting each pixel value from the maximum intensity value, replacing colors with their complementary colors. This transformation does not depend on $\epsilon$.
\end{itemize}

\begin{figure}[ht!]
    \centering
    \begin{subfigure}{0.4\columnwidth}
        \centering
        \includegraphics[width=\linewidth]{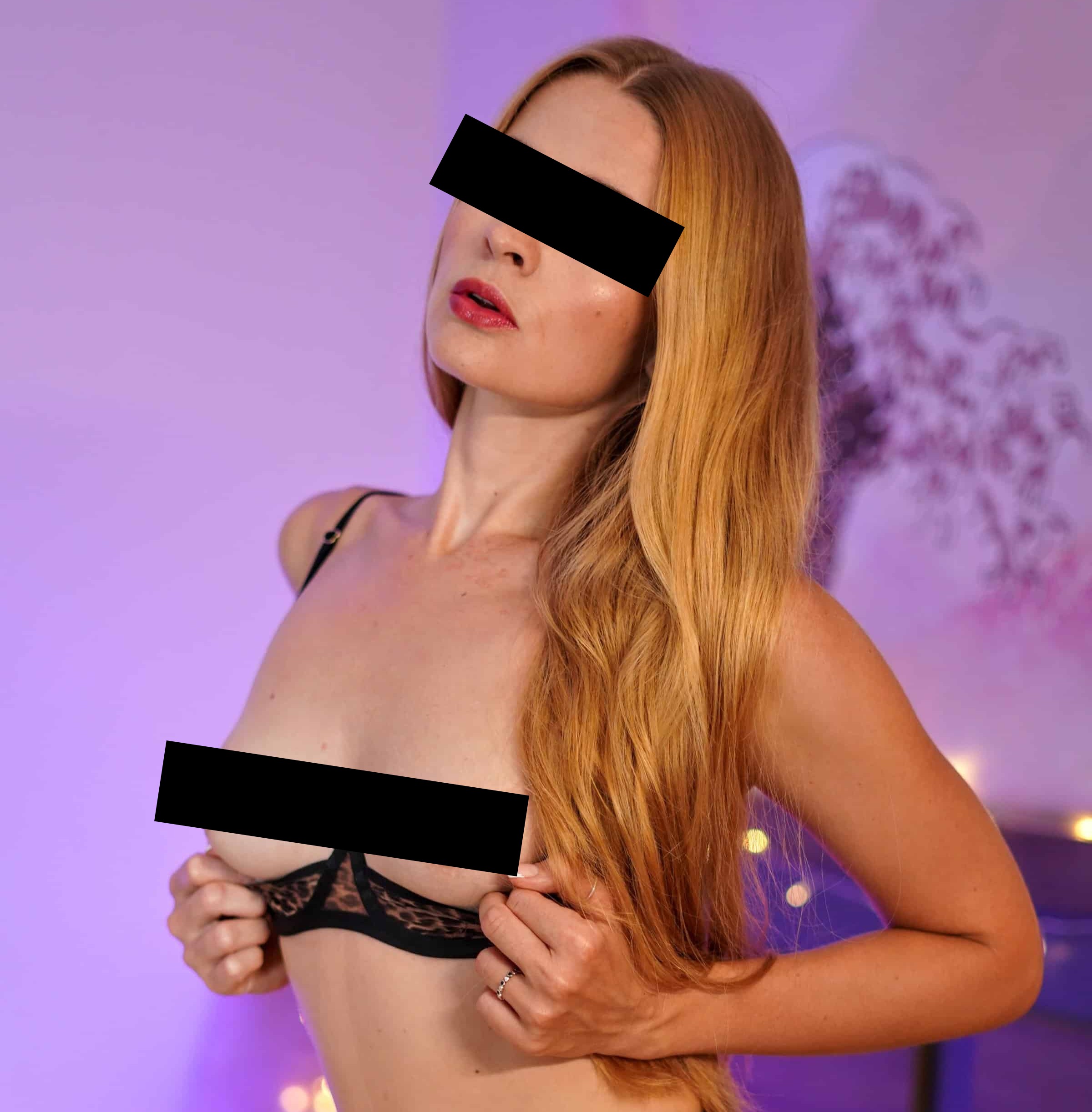}
        \caption{Original image}
        \label{fig:salt_pepper_original}
    \end{subfigure}
    \hfill
    \begin{subfigure}{0.4\columnwidth}
        \centering
        \includegraphics[width=\linewidth]{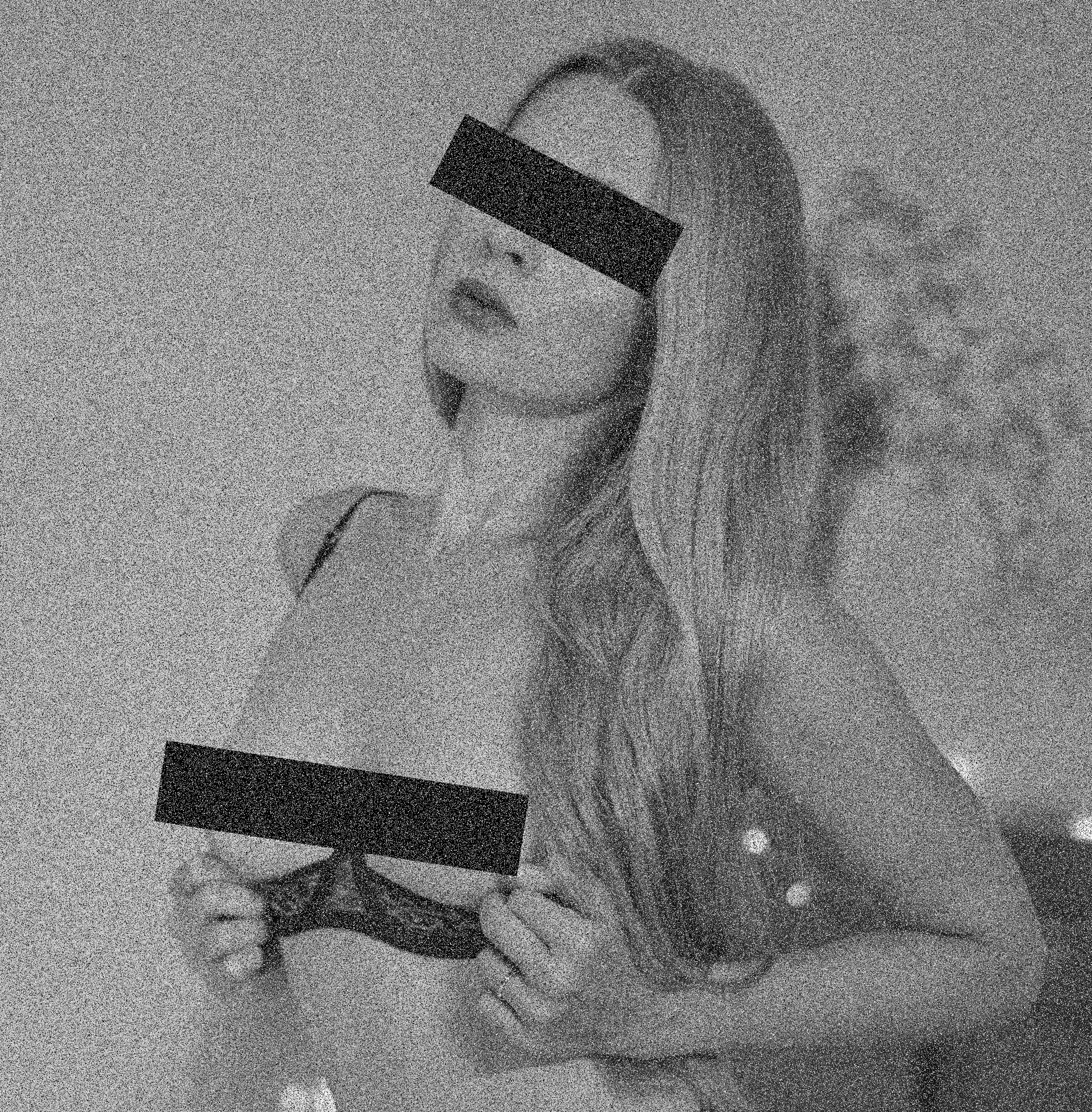}
        \caption{Salt and Pepper}
        \label{fig:salt_pepper_perturbed}
    \end{subfigure}
    \caption{Example of a Salt and Pepper transformation. Both images are censored solely for presentation purposes; all experiments were conducted on the original, uncensored dataset images.}
    \label{fig:salt_pepper_example}
\end{figure}

\noindent
Figure~\ref{fig:salt_pepper_example} shows an example of the Salt and Pepper transformation. While \texttt{omni-moderation} flags the original image as unsafe, the transformed image on the right is classified as safe, thereby bypassing the moderation decision.

\subsection{Datasets}
\label{sec:datasets}
For our evaluation, we use three datasets representing complementary forms of unsafe content: \bcircle{1} LSPD~\cite{phan2022lspd}, a large-scale dataset for pornographic-content detection; \bcircle{2} UnsafeBench~\cite{qu2025unsafebench}, a policy-oriented image-safety benchmark; and \bcircle{3} Hateful Memes~\cite{kiela2020hateful}, a multimodal benchmark in which harmful meaning may emerge from the interaction between images and embedded text. 

\noindent
\textbf{\bcircle{1} LSPD.}
The Large-Scale Pornographic Dataset (LSPD) contains more than \num{500000} pornographic and non-pornographic images and was designed for pornographic-content detection and classification~\cite{phan2022lspd}. Its relevance to multimodal-safety research is further supported by its use as the source of pornographic images in ToViLaG~\cite{wang2023tovilag}. We use LSPD to evaluate robustness on sexually explicit visual content.
Given the huge size of the dataset, we draw a statistically significant random sample rather than evaluating the entire dataset. Under a conservative population-proportion estimate, a 99\% confidence level and a 5\% margin of error require approximately 665 samples. We round this value up to 1000 images to obtain a larger and more convenient evaluation set.

\noindent
\textbf{\bcircle{2} UnsafeBench.}
UnsafeBench is a policy-oriented image-safety benchmark comprising \num{10146} valid images across 11 unsafe-content categories: deception, harassment, hate, illegal activity, political content, public or personal health, self-harm, sexual content, shocking content, spam, and violence~\cite{qu2025unsafebench}. Since only a subset of the categories supported by \texttt{omni-moderation} can be detected from image inputs~\cite{openai_moderation_docs}, we retain only the \emph{sexual}, \emph{violence}, and \emph{self-harm} categories. This prevents introducing a systematic bias into the robustness evaluation, since unflagged outputs for unsupported categories could otherwise be incorrectly attributed to robustness rather than to the model's documented lack of visual coverage.

\noindent
\textbf{\bcircle{3} Hateful Memes.}
Hateful Memes~\cite{kiela2020hateful} is a multimodal benchmark introduced by Meta as part of the Hateful Memes Challenge. It contains \num{11605} memes whose meaning may depend on the interaction between embedded text and visual content. This dataset complements LSPD and UnsafeBench by testing content whose moderation requires reading text rendered inside an image and relating it to the surrounding visual context. We submit each meme as a single image, without providing a separate textual transcription.

Our robustness analysis does not use the datasets' ground-truth annotations as the initial moderation labels. Prior work has indeed shown that UnsafeBench may contain "mislabeled" images and that its \textit{safe}/\textit{unsafe} annotations may not fully align with moderation systems~\cite{zeng2025shieldgemma}. Therefore, for each target system, we first query the clean image and treat the returned decision as the system-specific baseline. We then restrict the attack evaluation to images initially classified as unsafe and measure whether a transformation changes the decision to safe.
This design separates robustness from baseline classification accuracy and avoids conflating transformation-induced decision changes with discrepancies among dataset annotations, provider taxonomies, and moderation policies. 
\section{Experimental Results}
\label{sec:experimentalResults}
In this section, we present the experimental results organized according to the following three research questions (RQs):
\begin{description}
    \item[RQ1:] To what extent are commercial image-moderation services vulnerable to common image transformations?
    \item[RQ2:] How does moderation robustness vary across datasets representing different types of unsafe and multimodal content?
    \item[RQ3:] What is the minimum transformation intensity required to change an unsafe moderation decision into a safe one?
\end{description}

\begin{figure*}[ht]
    \centering
    \includegraphics[width=\linewidth]{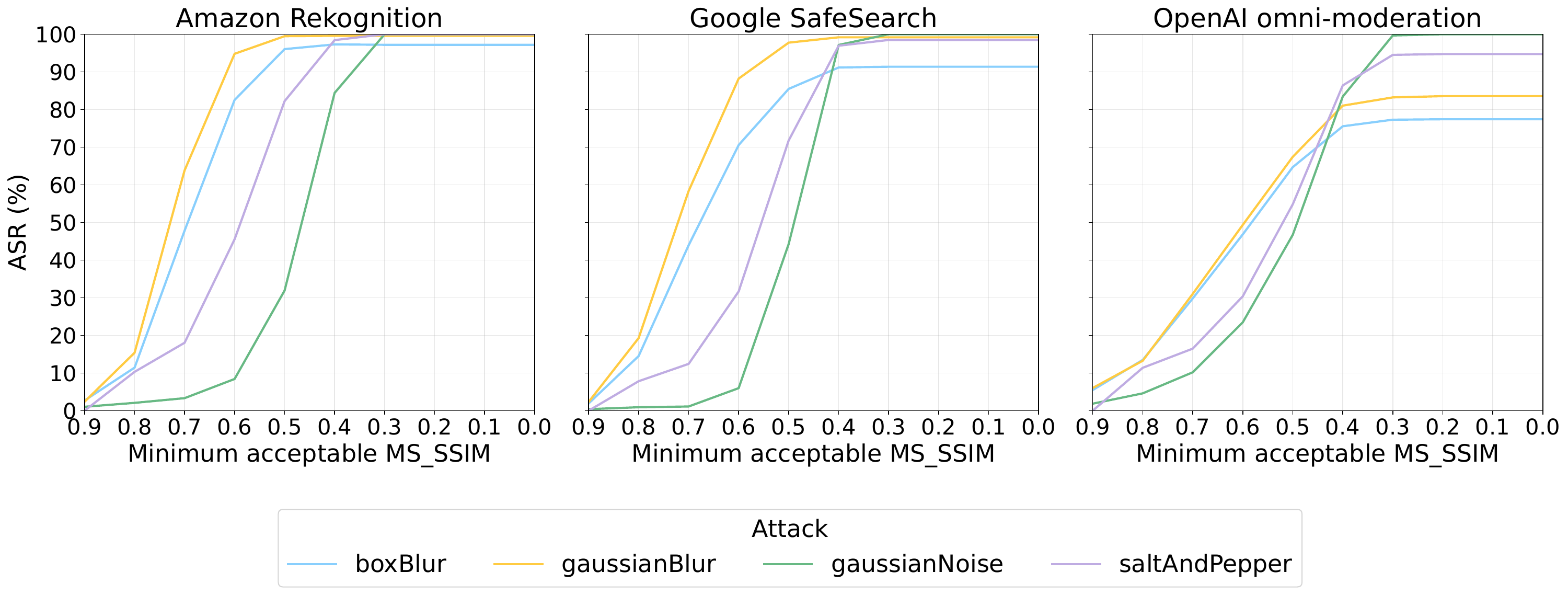}
    \caption{ASR of intensity-dependent transformations across different minimum acceptable MS-SSIM thresholds on LSPD.}
    \label{fig:rq1-intensity-dependent}
\end{figure*}

\subsection{RQ1: Robustness Across Commercial APIs}
\label{sec:rq1}
In RQ1, we evaluate whether simple, model-agnostic image transformations can bypass OpenAI \texttt{omni-moderation}, Amazon Rekognition, and Google Cloud SafeSearch. To enable a controlled cross-provider comparison, we focus on pornographic content, which is explicitly covered by all three services, and use the 1000-image LSPD sample~\cite{phan2022lspd} described in Section~\ref{sec:datasets}. Evaluating all providers across all datasets would introduce substantial financial and operational overhead. We therefore use LSPD as the common evaluation setting for RQ1, while broader cross-dataset and category-level analyses are examined in the subsequent RQs.

We define the Attack Success Rate (ASR) as the percentage of images initially classified as 'unsafe' that become classified as 'safe' after applying a transformation. In other words, an attack is successful when a moderation service flags the clean image as unsafe but accepts its transformed version as safe. More formally, let \(f(x)\) denote the binary moderation decision for an image \(x\), where the output is either \texttt{safe} or \texttt{unsafe}. Let \(T(x)\) denote the image obtained by applying transformation \(T\) to \(x\). We define the ASR of \(T\) as:

\[
\mathrm{ASR}(T)=
\frac{
\#\left\{x :
f(x)=\texttt{unsafe}
\ \land\
f(T(x))=\texttt{safe}
\right\}
}{
\#\left\{x :
f(x)=\texttt{unsafe}
\right\}
}.
\]

\noindent
Since Amazon Rekognition and Google Cloud SafeSearch do not directly return binary decisions, we derive \texttt{safe}/\texttt{unsafe} labels from their outputs. For Amazon Rekognition, we classify an image as \texttt{unsafe} when at least one moderation label is returned. For Google Cloud SafeSearch, we classify an image as \texttt{unsafe} when at least one considered category is assigned a likelihood of \texttt{LIKELY} or \texttt{VERY\_LIKELY}.

As described in Section~\ref{sec:adversarialModifications}, we evaluate both \wcircle{1} \textit{one-shot transformations} and \wcircle{2} \textit{intensity-dependent transformations}. One-shot transformations are applied once, and therefore their ASR is computed directly. For intensity-dependent transformations, the scale of \(\epsilon\) varies across attacks, so fixed values are not directly comparable. Moreover, \(\epsilon\) could be increased until the image is severely degraded or effectively destroyed, making any resulting moderation-label change uninformative. 
We therefore compare intensity-dependent attacks under a common image-preservation constraint based on the Multi-Scale Structural Similarity Index Measure (MS-SSIM)~\cite{wang2003multiscale}. Given a minimum acceptable MS-SSIM threshold \(\tau\), for each image and attack, we retain the perturbed version generated with the largest tested intensity \(\epsilon\) whose MS-SSIM with respect to the clean image is greater than or equal to \(\tau\). The ASR is then computed using these selected perturbations. 
Table~\ref{tab:rq1-one-shot-transformations} reports the ASR of the one-shot transformations, while Figure~\ref{fig:rq1-intensity-dependent} reports the ASR of the intensity-dependent transformations across the considered minimum MS-SSIM thresholds \(\tau\).

\begin{table}[ht]
\centering
\caption{ASR of one-shot transformations on the LSPD pornographic subset.}
\label{tab:rq1-one-shot-transformations}
\small
\begin{adjustbox}{width=0.8\columnwidth}
\begin{tabular}{lccc}
\toprule
\textbf{Transformation}
& \textbf{Amazon}
& \textbf{Google}
& \textbf{OpenAI} \\
\midrule
Greyscale       & 1.77\%  & 1.12\% & 4.17\% \\
Invert Color    & 43.97\% & 6.29\% & 8.11\% \\
Split-Merge RGB & 3.12\%  & 0.81\% & 1.32\% \\
\bottomrule
\end{tabular}
\end{adjustbox}
\end{table}

Even one-shot transformations can bypass all three services. In particular, color inversion reaches an ASR of \(43.97\%\) for Amazon Rekognition, \(6.29\%\) for Google Cloud SafeSearch, and \(8.11\%\) for OpenAI \texttt{omni-moderation}, while greyscale and Split-Merge RGB also produce non-zero bypass rates. Intensity-dependent transformations are broadly effective across all providers. As expected, relaxing the MS-SSIM constraint increases the ASR because stronger transformations become admissible. However, substantial bypass rates also emerge before the similarity constraint is fully relaxed. For instance, at \(\tau=0.6\), Salt and Pepper reaches an ASR of \(45.53\%\) against Amazon Rekognition and above \(30\%\) against both Google Cloud SafeSearch (\(31.68\%\)) and OpenAI \texttt{omni-moderation} (\(30.37\%\)). This indicates that successful bypasses do not require completely destroying the original image. These weaknesses have direct practical implications for platforms relying on such services: a malicious user could apply inexpensive transformations to pornographic images before uploading them, increasing the likelihood that unsafe content passes automated moderation and reaches other users.

\begin{answer}
\textbf{Answer to RQ1:} All three commercial image-moderation services are vulnerable to both one-shot and intensity-dependent transformations, with successful bypasses also occurring under relatively high image-similarity constraints. This creates practical risks for platforms that rely on such APIs.    
\end{answer}

\subsection{RQ2: Robustness Across Datasets and Content Categories}
\label{sec:rq2}

While RQ1 provided an initial cross-provider comparison focused on pornographic content, RQ2 broadens the analysis to different forms of unsafe and multimodal content. This broader evaluation requires a substantially larger number of queries across datasets and transformations; we therefore focus on OpenAI \texttt{omni-moderation}, whose accessibility makes such a large-scale analysis operationally feasible. We investigate whether its robustness varies across the LSPD, UnsafeBench, and Hateful Memes datasets described in Section~\ref{sec:datasets}. 

\begin{figure*}[ht]
    \centering
    \includegraphics[width=\linewidth]{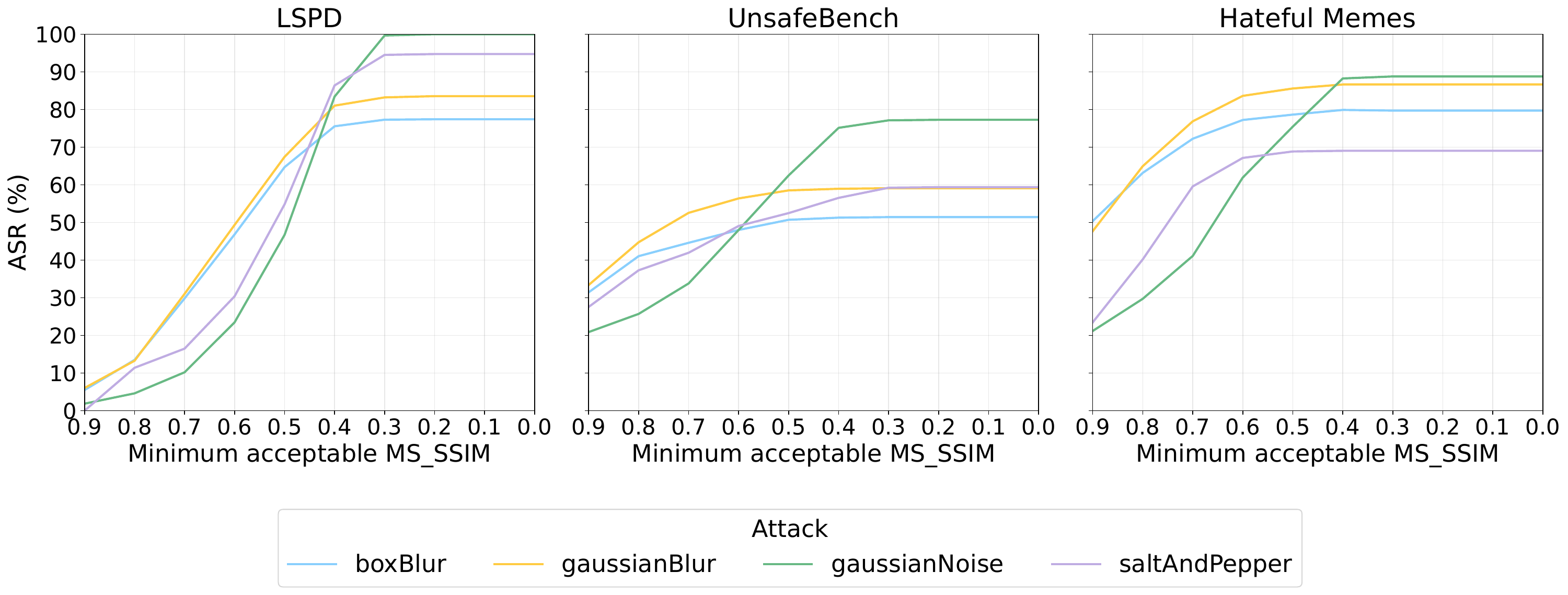}
    \caption{ASR of intensity-dependent transformations across the three datasets and different minimum acceptable MS-SSIM thresholds using OpenAI \texttt{omni-moderation}}
    \label{fig:rq2-intensity-dependent}
\end{figure*}

We use the ASR defined in Section~\ref{sec:rq1} and follow the same evaluation protocol. One-shot transformations are applied once and evaluated directly, while intensity-dependent transformations are assessed under the MS-SSIM constraint introduced in RQ1. Table~\ref{tab:rq2-one-shot-transformations} reports the ASR of one-shot transformations across the three datasets, while Figure~\ref{fig:rq2-intensity-dependent} reports the results for intensity-dependent transformations across the considered MS-SSIM thresholds.

\begin{table}[ht]
\centering
\caption{ASR of one-shot transformations across the three datasets using OpenAI \texttt{omni-moderation}.}
\label{tab:rq2-one-shot-transformations}
\small
\begin{adjustbox}{width=0.8\columnwidth}
\begin{tabular}{lccc}
\toprule
\textbf{Transformation}
& \textbf{LSPD}
& \textbf{UnsafeBench}
& \textbf{Hateful Memes} \\
\midrule
Greyscale       & 4.17\% & 16.34\% & 13.61\% \\
Invert Color    & 8.11\% & 34.66\% & 21.47\% \\
Split-Merge RGB & 1.32\% & 16.90\% & 14.14\% \\
\bottomrule
\end{tabular}
\end{adjustbox}
\end{table}

It can be observed that one-shot transformations remain effective across all three datasets. UnsafeBench exhibits the highest ASRs within the supported-category subset, with color inversion reaching \(34.66\%\), while Hateful Memes also shows substantial bypass rates despite requiring joint interpretation of embedded text and visual content.
A similar pattern emerges for intensity-dependent transformations. Relaxing the MS-SSIM constraint increases the ASR across all datasets, but the magnitude differs considerably. Hateful Memes is already highly vulnerable at relatively strict thresholds, while UnsafeBench shows more moderate but consistent bypass rates. LSPD, which contains homogeneous pornographic imagery, starts with lower ASR values and increases more sharply as stronger transformations are allowed.
Overall, the results indicate that robustness depends on the type and modality of unsafe content. Simple image transformations affect both purely visual pornographic content and multimodal hateful memes in which meaning emerges from the interaction between text and image.

\begin{figure}[h!]
    \centering
    \includegraphics[width=\columnwidth]{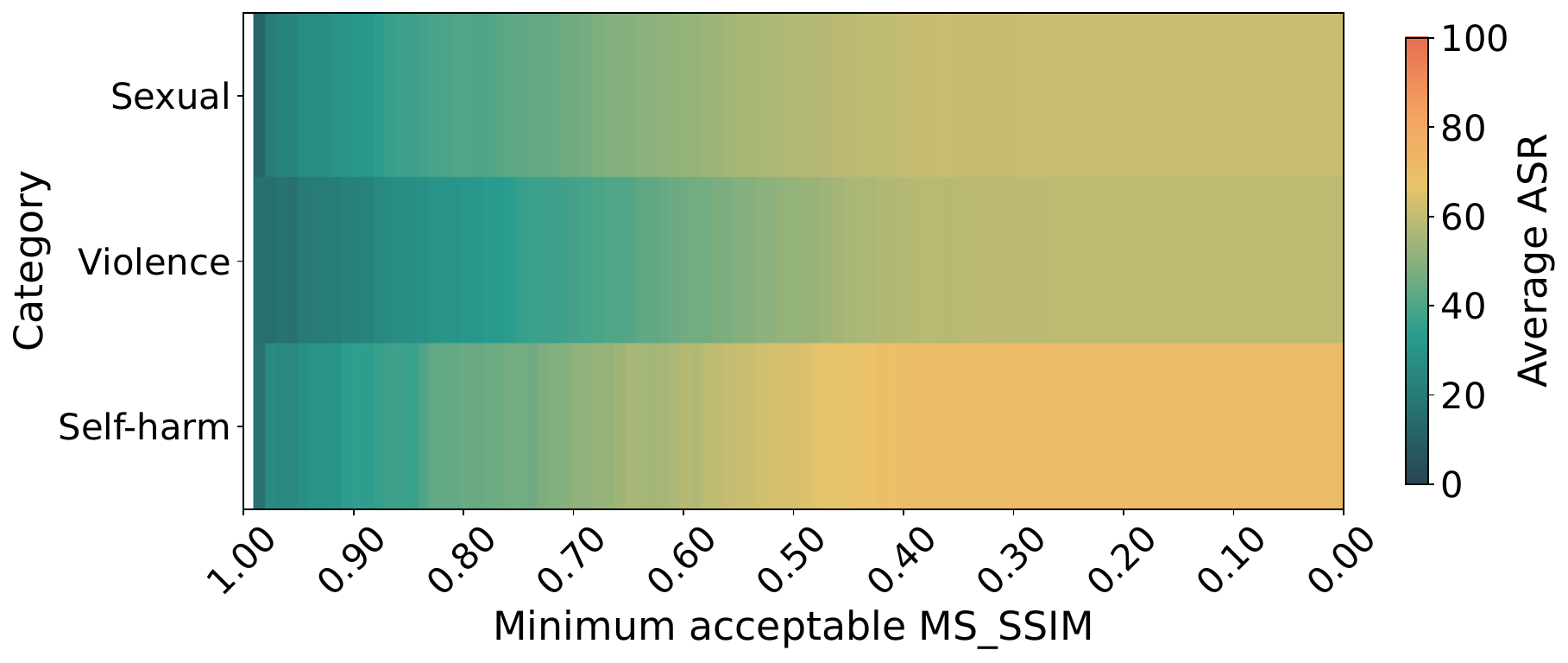}
    \caption{Average category-wise ASR on the UnsafeBench subset across MS-SSIM thresholds.}
    \label{fig:rq2b_category}
\end{figure}

\noindent
\textbf{Intra-dataset category comparison.}
The first part of RQ2 compared datasets collected from different sources and representing different types of unsafe content. We further examine whether robustness also varies within the same dataset across the three UnsafeBench categories considered in our study: sexual content, violence, and self-harm. Figure~\ref{fig:rq2b_category} reports, for each category and MS-SSIM threshold, the ASR averaged across all tested transformations.

The results show that self-harm is consistently the most vulnerable category, followed by sexual content, while violence exhibits the lowest ASR across most thresholds. Thus, the robustness of \texttt{omni-moderation} varies not only across datasets, but also across unsafe-content categories.

\begin{answer}
\textbf{Answer to RQ2:} The robustness of \texttt{omni-moderation} varies across datasets and content categories. Simple transformations affect both purely visual and multimodal inputs, while self-harm content is more vulnerable than sexual and violent content within UnsafeBench.   
\end{answer}

\subsection{RQ3: Minimum Transformation Intensity}
\label{sec:rq3}
In RQ1, we evaluated transformation effectiveness under a shared perceptual-similarity constraint, selecting for each transformation the strongest intensity that preserved a given MS-SSIM threshold. RQ3 addresses a complementary question: rather than fixing image similarity and measuring the resulting ASR, we identify the minimum tested transformation intensity required to change the decision of OpenAI \texttt{omni-moderation}. We reuse the same 1000-image LSPD sample to enable direct comparison with the OpenAI results in RQ1 while controlling for dataset and content differences. The analysis is restricted to intensity-dependent transformations, since a minimum successful intensity is not defined for one-shot transformations without an ordered strength scale.

We define the \emph{minimum successful transformation intensity} for an image \(x\) and transformation \(T\) as the smallest tested intensity \(\epsilon\) that changes the moderation decision from \texttt{unsafe} to \texttt{safe}. Formally,
\[
\epsilon^{*}(x,T)=
\min \left\{
\epsilon :
f(x)=\texttt{unsafe}
\ \land\
f(T_{\epsilon}(x))=\texttt{safe}
\right\},
\]

\noindent
where \(f(\cdot)\) denotes the binary decision returned by \texttt{omni-moderation}, and \(T_{\epsilon}(x)\) is the image obtained by applying transformation \(T\) to \(x\) at intensity \(\epsilon\).

\begin{figure}[ht]
    \centering
    \includegraphics[width=\columnwidth]{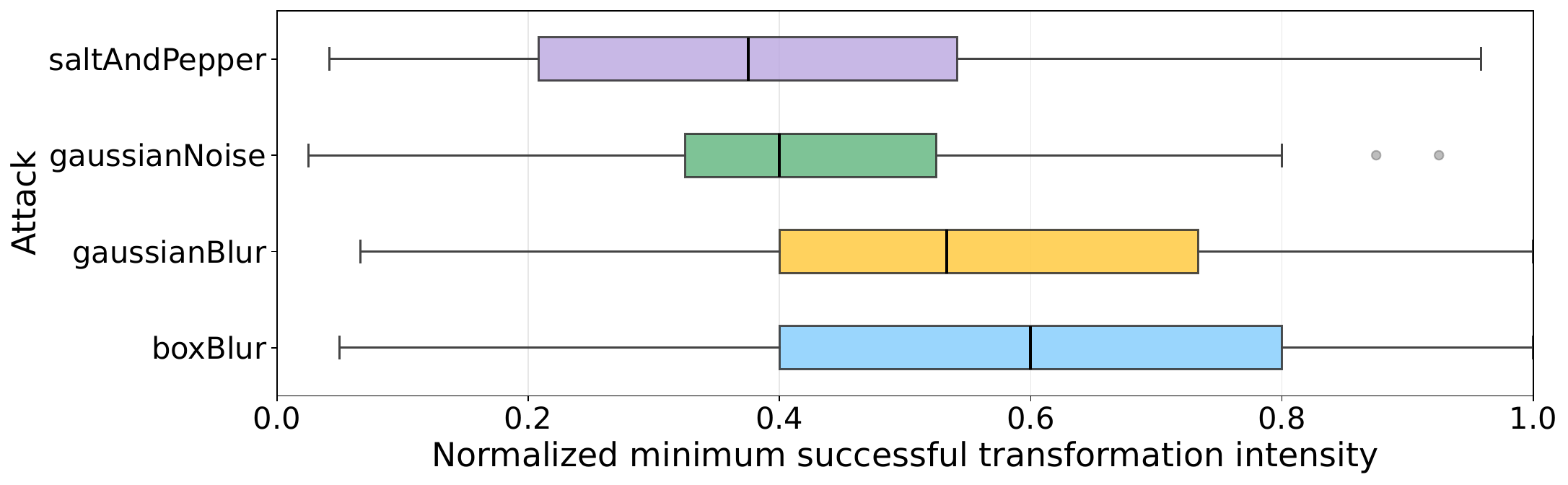}
    \caption{Distribution of the normalized minimum successful transformation intensity on the LSPD pornographic subset.}
    \label{fig:rq3}
\end{figure}

Before analyzing the minimum successful intensity, we exclude images that remain \texttt{unsafe} across all tested intensities, since no minimum successful intensity is defined for them. 
We then analyze only the images that undergo at least one label change, considering the intensity required for the first \texttt{unsafe}-to-\texttt{safe} transition.  Figure~\ref{fig:rq3} reports the distribution of the minimum successful transformation intensity. Since intensity values have different meanings and scales across transformation families, each value is normalized by the maximum tested intensity for the corresponding transformation.

The distributions show that relatively low transformation intensities are sufficient to change the label for many images, particularly for Salt and Pepper and Gaussian Noise, whose median minimum intensities are around 0.4 of the tested range. Values are distributed across a broad portion of the range, with some images changing label at very low intensities and others requiring substantially stronger transformations. The wide interquartile ranges and whiskers, therefore, indicate considerable image-level variability. Unlike RQ1, this analysis imposes no perceptual-similarity constraint: it identifies only the first tested intensity at which the moderation decision changes from \texttt{unsafe} to \texttt{safe}.

\begin{answer}
\textbf{Answer to RQ3:} OpenAI \texttt{omni-moderation} can often be destabilized well before the maximum tested transformation intensity. Although the required strength varies across transformations and images, many pornographic images undergo an \texttt{unsafe}-to-\texttt{safe} label change at low or moderate normalized intensities.    
\end{answer}

\section{Discussion}
\label{sec:discussion}
Our findings reveal a marked asymmetry between attacker effort and defender burden. Simple, model-agnostic image transformations can induce \texttt{unsafe}-to-\texttt{safe} decision changes without model access, gradients, training data, or surrogate models. They can be implemented with widely available tools and applied automatically at scale, keeping the cost of evasion low. Defenders, by contrast, may require repeated moderation checks, transformation-aware preprocessing, additional classifiers, or human review, increasing computational and operational overhead. This imbalance is particularly concerning when moderation APIs serve as the first or only line of defense for user-generated content. 
The practical implications extend beyond a single provider, with vulnerability affecting all three evaluated commercial services. These systems should therefore not be treated as standalone security boundaries. Instead, they should be integrated into layered moderation pipelines that combine multiple signals, periodically test robustness under realistic transformations, and escalate uncertain or high-risk cases to additional automated or human review.
\section{Limitations}
\label{sec:limitations}
In this section, we summarize the main limitations and threats to validity of our study.

\noindent
\textbf{Experimental Scale and Cost.}
The large number of combinations across systems, datasets, transformations, and intensity levels requires a substantial volume of moderation queries. Evaluating every system across all such combinations would therefore introduce considerable financial and operational overhead. To mitigate this limitation while still enabling a fair cross-provider comparison, RQ1 evaluates all three commercial services under the same controlled setting. The broader analyses in RQ2 and RQ3 instead focus on OpenAI \texttt{omni-moderation}, which is freely accessible, making evaluation at this scale feasible.

\noindent
\textbf{Generalization Across Moderation Systems.}
Although RQ1 shows that all three evaluated commercial services are vulnerable to simple image transformations, this does not imply that every moderation system fails in the same way or to the same extent. Open-weight models, such as LLaVAGuard~\cite{helff2024llavaguard} and ShieldGemma 2~\cite{zeng2025shieldgemma}, are relevant additional targets but require local deployment and suitable hardware. We focus on commercial APIs because they are easier to integrate and more accessible to developers and platforms without dedicated moderation infrastructure. Evaluating open-weight models remains an important direction for future work.

\noindent
\textbf{Dataset Generalization.}
Our findings depend on the selected benchmarks, which may not capture the full diversity of real-world unsafe content or align perfectly with each provider's policy taxonomy. We mitigate this limitation by evaluating well-established datasets from the literature that cover different types of unsafe content. Nevertheless, our conclusions remain limited to the evaluated images, categories, and transformations and should not be generalized to all unsafe content.
\section{Conclusions}
\label{sec:conclusions}
This paper presented a large-scale black-box robustness audit of three commercial image-moderation APIs under simple, model-agnostic transformations. We first compared OpenAI \texttt{omni-moderation}, Amazon Rekognition, and Google Cloud SafeSearch, and then analyzed \texttt{omni-moderation} more broadly across datasets, harm categories, and transformation intensities. Our results show that low-cost transformations can induce substantial \texttt{unsafe}-to-\texttt{safe} decision changes without model access, gradients, surrogate models, or optimized adversarial examples. These findings suggest that commercial moderation APIs should not be treated as standalone security boundaries. Future work may further extend the evaluation to additional commercial and open-weight systems.

\bibliography{aaai2027}

@inproceedings{alecci2023your,
  title={Your attack is too dumb: Formalizing attacker scenarios for adversarial transferability},
  author={Alecci, Marco and Conti, Mauro and Marchiori, Francesco and Martinelli, Luca and Pajola, Luca},
  booktitle={Proceedings of the 26th international symposium on research in attacks, intrusions and defenses},
  pages={315--329},
  year={2023}
}

@misc{openai2024omnimoderation,
  author       = {{OpenAI}},
  title        = {Upgrading the Moderation {API} with Our New Multimodal Moderation Model},
  year         = {2024},
  howpublished = {\url{https://openai.com/index/upgrading-the-moderation-api-with-our-new-multimodal-moderation-model/}},
  note         = {September 26, 2024. Accessed July 27, 2026}
}

@inproceedings{markov2023holistic,
  title={A holistic approach to undesired content detection in the real world},
  author={Markov, Todor and Zhang, Chong and Agarwal, Sandhini and Nekoul, Florentine Eloundou and Lee, Theodore and Adler, Steven and Jiang, Angela and Weng, Lilian},
  booktitle={Proceedings of the AAAI conference on artificial intelligence},
  volume={37},
  pages={15009--15018},
  year={2023}
}

@article{szegedy2013intriguing,
  title={Intriguing properties of neural networks},
  author={Szegedy, Christian and Zaremba, Wojciech and Sutskever, Ilya and Bruna, Joan and Erhan, Dumitru and Goodfellow, Ian and Fergus, Rob},
  journal={arXiv preprint arXiv:1312.6199},
  year={2013}
}

@inproceedings{goodfellow2014explaining,
  title     = {Explaining and Harnessing Adversarial Examples},
  author    = {Goodfellow, Ian J. and Shlens, Jonathon and Szegedy, Christian},
  booktitle = {International Conference on Learning Representations},
  year      = {2015}
}

@inproceedings{madry2017towards,
  title     = {Towards Deep Learning Models Resistant to Adversarial Attacks},
  author    = {Madry, Aleksander and Makelov, Aleksandar and Schmidt, Ludwig and Tsipras, Dimitris and Vladu, Adrian},
  booktitle = {International Conference on Learning Representations},
  year      = {2018}
}

@inproceedings{papernot2017practical,
  title={Practical black-box attacks against machine learning},
  author={Papernot, Nicolas and McDaniel, Patrick and Goodfellow, Ian and Jha, Somesh and Celik, Z Berkay and Swami, Ananthram},
  booktitle={Proceedings of the 2017 ACM on Asia conference on computer and communications security},
  pages={506--519},
  year={2017}
}

@inproceedings{marchiori2025dumb,
  title={{DUMB} and {DUMBer}: Is Adversarial Training Worth It in the Real World?},
  author={Marchiori, Francesco and Alecci, Marco and Pajola, Luca and Conti, Mauro},
  booktitle={European Symposium on Research in Computer Security},
  pages={228--248},
  year={2025},
  organization={Springer}
}

@inproceedings{hendrycks2019benchmarking,
  title     = {Benchmarking Neural Network Robustness to Common Corruptions and Perturbations},
  author    = {Hendrycks, Dan and Dietterich, Thomas},
  booktitle = {International Conference on Learning Representations},
  year      = {2019}
}

@inproceedings{tian2018deeptest,
  title={Deeptest: Automated testing of deep-neural-network-driven autonomous cars},
  author={Tian, Yuchi and Pei, Kexin and Jana, Suman and Ray, Baishakhi},
  booktitle={Proceedings of the 40th international conference on software engineering},
  pages={303--314},
  year={2018}
}

@inproceedings{hendrycks2020augmix,
  title={Augmix: A simple method to improve robustness and uncertainty under data shift},
  author={Hendrycks, Dan and Mu, Norman and Cubuk, Ekin Dogus and Zoph, Barret and Gilmer, Justin and Lakshminarayanan, Balaji},
  booktitle={International conference on learning representations},
  volume={2},
  pages={6},
  year={2020}
}

@inproceedings{hartmann2025lost,
  title={Lost in moderation: How commercial content moderation apis over-and under-moderate group-targeted hate speech and linguistic variations},
  author={Hartmann, David and Oueslati, Amin and Staufer, Dimitri and Pohlmann, Lena and Munzert, Simon and Heuer, Hendrik},
  booktitle={Proceedings of the 2025 CHI Conference on Human Factors in Computing Systems},
  pages={1--26},
  year={2025}
}

@inproceedings{qu2025unsafebench,
  title={{UnsafeBench}: Benchmarking image safety classifiers on real-world and ai-generated images},
  author={Qu, Yiting and Shen, Xinyue and Wu, Yixin and Backes, Michael and Zannettou, Savvas and Zhang, Yang},
  booktitle={Proceedings of the 2025 ACM SIGSAC Conference on Computer and Communications Security},
  pages={3221--3235},
  year={2025}
}

@article{zeng2025shieldgemma,
  title={ShieldGemma 2: Robust and tractable image content moderation},
  author={Zeng, Wenjun and Kurniawan, Dana and Mullins, Ryan and Liu, Yuchi and Saha, Tamoghna and Ike-Njoku, Dirichi and Gu, Jindong and Song, Yiwen and Xu, Cai and Zhou, Jingjing and others},
  journal={arXiv preprint arXiv:2504.01081},
  year={2025}
}

@article{yuan2026promptguard,
  title={Promptguard: Soft prompt-guided unsafe content moderation for text-to-image models},
  author={Yuan, Lingzhi and Li, Xinfeng and Xu, Chejian and Tao, Guanhong and Jia, Xiaojun and Huang, Yihao and Dong, Wei and Liu, Yang and Li, Bo},
  journal={IEEE Transactions on Information Forensics and Security},
  year={2026},
  publisher={IEEE}
}

@article{yang2026eyes,
  title={What the Eyes See, the LLMs Miss: Exploiting Human Perception for Adversarial Text Attacks},
  author={Yang, Qin and Malloy, Lu and Lee, Joshua and Chang, Xiaohan and Mohammady, Meisam and Kim, Doowon and Hong, Yuan},
  journal={arXiv preprint arXiv:2606.09700},
  year={2026}
}

@inproceedings{li2026making,
  title={Making mllms blind: Adversarial smuggling attacks in mllm content moderation},
  author={Li, Zhiheng and Ma, Zongyang and Pan, Yuntong and Zhang, Ziqi and Lv, Xiaolei and Li, Bo and Gao, Jun and Zhang, Jianing and Yuan, Chunfeng and Li, Bing and others},
  booktitle={Findings of the Association for Computational Linguistics: ACL 2026},
  pages={20142--20161},
  year={2026}
}

@inproceedings{grondahl2018all,
  title={All you need is "love" evading hate speech detection},
  author={Gr{\"o}ndahl, Tommi and Pajola, Luca and Juuti, Mika and Conti, Mauro and Asokan, N},
  booktitle={Proceedings of the 11th ACM workshop on artificial intelligence and security},
  pages={2--12},
  year={2018}
}

@book{gillespie2018custodians,
  title={Custodians of the Internet: Platforms, content moderation, and the hidden decisions that shape social media},
  author={Gillespie, Tarleton},
  year={2018},
  publisher={Yale University Press}
}

@book{roberts2019behind,
  title={Behind the Screen: Content Moderation in the Shadows of Social Media},
  author={Roberts, Sarah T.},
  year={2019},
  publisher={Yale University Press}
}

@misc{perspectiveapi2026,
  author       = {{Jigsaw}},
  title        = {Perspective {API}},
  year         = {n.d.},
  howpublished = {\url{https://www.perspectiveapi.com/}},
  note         = {Accessed July 27, 2026}
}

@article{malik2025deep,
  title={Deep learning for hate speech detection: a comparative study},
  author={Malik, Jitendra Singh and Qiao, Hezhe and Pang, Guansong and van den Hengel, Anton},
  journal={International Journal of Data Science and Analytics},
  volume={20},
  number={4},
  pages={3053--3068},
  year={2025},
  publisher={Springer}
}

@misc{googlecloudenaturallanguageapi,
  author       = {{Google Cloud}},
  title        = {Cloud Natural Language Documentation},
  year         = {2026},
  howpublished = {\url{https://cloud.google.com/natural-language/docs}},
  note         = {Accessed July 27, 2026}
}

@inproceedings{caselli2021hatebert,
  title={HateBERT: Retraining BERT for abusive language detection in English},
  author={Caselli, Tommaso and Basile, Valerio and Mitrovi{\'c}, Jelena and Granitzer, Michael},
  booktitle={Proceedings of the 5th Workshop on Online Abuse and Harms (WOAH 2021)},
  pages={17--25},
  year={2021}
}

@inproceedings{sarkar2021fbert,
  title={fBERT: A neural transformer for identifying offensive content},
  author={Sarkar, Diptanu and Zampieri, Marcos and Ranasinghe, Tharindu and Ororbia, Alexander},
  booktitle={Findings of the association for computational linguistics: EMNLP 2021},
  pages={1792--1798},
  year={2021}
}

@inproceedings{pandey2021device,
  title={On-device content moderation},
  author={Pandey, Anchal and Moharana, Sukumar and Mohanty, Debi Prasanna and Panwar, Archit and Agarwal, Dewang and Thota, Siva Prasad},
  booktitle={2021 International Joint Conference on Neural Networks (IJCNN)},
  pages={1--7},
  year={2021},
  organization={IEEE}
}

@misc{schuhmann_clip_nsfw_detector_2022,
  author       = {Schuhmann, Christoph and {LAION-AI}},
  title        = {{CLIP}-Based {NSFW} Detector},
  year         = {2022},
  howpublished = {\url{https://github.com/LAION-AI/CLIP-based-NSFW-Detector}},
  note         = {Accessed July 27, 2026}
}

@misc{yahoo_open_nsfw,
  author       = {{Yahoo}},
  title        = {Open {NSFW} Model},
  year         = {2016},
  howpublished = {\url{https://github.com/yahoo/open_nsfw}},
  note         = {Archived October 18, 2019. Accessed July 27, 2026}
}

@misc{microsoft_azure_ai_content_safety_overview,
  author       = {{Microsoft}},
  title        = {What Is Azure {AI} Content Safety?},
  year         = {2026},
  howpublished = {\url{https://learn.microsoft.com/en-us/azure/ai-services/content-safety/overview}},
  note         = {Accessed July 27, 2026}
}

@article{kiela2020hateful,
  title={The Hateful Memes Challenge: Detecting hate speech in multimodal memes},
  author={Kiela, Douwe and Firooz, Hamed and Mohan, Aravind and Goswami, Vedanuj and Singh, Amanpreet and Ringshia, Pratik and Testuggine, Davide},
  journal={Advances in neural information processing systems},
  volume={33},
  pages={2611--2624},
  year={2020}
}

@article{das2020detecting,
  title={Detecting hate speech in multi-modal memes},
  author={Das, Abhishek and Wahi, Japsimar Singh and Li, Siyao},
  journal={arXiv preprint arXiv:2012.14891},
  year={2020}
}

@misc{mistral2026moderation,
  author       = {{Mistral AI}},
  title        = {Moderation and Guardrailing},
  year         = {2026},
  howpublished = {\url{https://docs.mistral.ai/capabilities/guardrailing/}},
  note         = {Accessed July 27, 2026}
}

@article{inan2023llamaguard,
  title={Llama guard: Llm-based input-output safeguard for human-ai conversations},
  author={Inan, Hakan and Upasani, Kartikeya and Chi, Jianfeng and Rungta, Rashi and Iyer, Krithika and Mao, Yuning and Tontchev, Michael and Hu, Qing and Fuller, Brian and Testuggine, Davide and others},
  journal={arXiv preprint arXiv:2312.06674},
  year={2023}
}

@article{zeng2024shieldgemma,
  title={Shieldgemma: Generative ai content moderation based on gemma},
  author={Zeng, Wenjun and Liu, Yuchi and Mullins, Ryan and Peran, Ludovic and Fernandez, Joe and Harkous, Hamza and Narasimhan, Karthik and Proud, Drew and Kumar, Piyush and Radharapu, Bhaktipriya and others},
  journal={arXiv preprint arXiv:2407.21772},
  year={2024}
}

@inproceedings{wang2023tovilag,
  title={ToViLaG: Your visual-language generative model is also an evildoer},
  author={Wang, Xinpeng and Yi, Xiaoyuan and Jiang, Han and Zhou, Shanlin and Wei, Zhihua and Xie, Xing},
  booktitle={Proceedings of the 2023 Conference on Empirical Methods in Natural Language Processing},
  pages={3508--3533},
  year={2023}
}

@article{phan2022lspd,
  title={LSPD: A large-scale pornographic dataset for detection and classification},
  author={Phan, Dinh Duy and Nguyen, Thanh Thien and Nguyen, Quang Huy and Tran, Hoang Loc and Nguyen, Khac Ngoc Khoi and Vu, Duc Lung},
  journal={International Journal of Intelligent Engineering and Systems},
  volume={15},
  number={1},
  year={2022}
}

@misc{horwitz2024instagram,
  author       = {Horwitz, Jeff},
  title        = {Instagram Recommends Sexual Videos to Accounts for 13-Year-Olds, Tests Show},
  year         = {2024},
  howpublished = {\url{https://www.wsj.com/tech/instagram-recommends-sexual-videos-to-accounts-for-13-year-olds-tests-show-b6123c65}},
  note         = {June 20, 2024. Accessed July 27, 2026}
}

@misc{adams2021sexualimages,
  author       = {Adams, Richard},
  title        = {Three in Four Girls Have Been Sent Sexual Images via Apps, Report Finds},
  year         = {2021},
  howpublished = {\url{https://www.theguardian.com/media/2021/dec/06/three-in-four-girls-have-been-sent-sexual-images-via-apps-report-finds}},
  note         = {December 6, 2021. Accessed July 27, 2026}
}

@misc{das2022instagram,
  author       = {Das, Shanti},
  title        = {Instagram under Fire over Sexualised Child Images},
  year         = {2022},
  howpublished = {\url{https://www.theguardian.com/society/2022/apr/17/instagram-under-fire-over-sexualised-child-images}},
  note         = {April 17, 2022. Accessed July 27, 2026}
}

@misc{lomas2023metachildprotection,
  author       = {Lomas, Natasha},
  title        = {Meta Warned It Faces ``Heavy Sanctions'' in the {EU} If It Fails to Fix Child Protection Problems},
  year         = {2023},
  howpublished = {\url{https://techcrunch.com/2023/06/08/meta-child-protection-dsa-warning/}},
  note         = {June 8, 2023. Accessed July 27, 2026}
}

@inproceedings{wang2003multiscale,
  title        = {Multiscale Structural Similarity for Image Quality Assessment},
  author       = {Wang, Zhou and Simoncelli, Eero P. and Bovik, Alan C.},
  booktitle    = {The Thirty-Seventh Asilomar Conference on Signals, Systems \& Computers},
  volume       = {2},
  pages        = {1398--1402},
  year         = {2003},
  organization = {IEEE}
}

@misc{openai_moderation_docs,
  author       = {{OpenAI}},
  title        = {Moderations},
  year         = {2026},
  howpublished = {\url{https://platform.openai.com/docs/api-reference/moderations}},
  note         = {Accessed July 27, 2026}
}

@misc{aws_rekognition_moderation,
  author       = {{Amazon Web Services}},
  title        = {Using the Image and Video Moderation {APIs}},
  year         = {2026},
  howpublished = {\url{https://docs.aws.amazon.com/rekognition/latest/dg/moderation-api.html}},
  note         = {Accessed July 27, 2026}
}

@misc{googlecloud_vision_safesearch_2026,
  author       = {{Google Cloud}},
  title        = {Detect Explicit Content ({SafeSearch})},
  year         = {2026},
  howpublished = {\url{https://cloud.google.com/vision/docs/detecting-safe-search}},
  note         = {Accessed July 27, 2026}
}

@article{helff2024llavaguard,
  title={LlavaGuard: An open vlm-based framework for safeguarding vision datasets and models},
  author={Helff, Lukas and Friedrich, Felix and Brack, Manuel and Kersting, Kristian and Schramowski, Patrick},
  journal={arXiv preprint arXiv:2406.05113},
  year={2024}
}

\appendix
\section{Appendix}

\subsection{Baseline Experimental Results}
\label{app:baselineEvaluation}

Table~\ref{tab:baseline_evaluation} reports the baseline moderation decisions returned for the clean images in each evaluation dataset. As discussed in the main paper, we use these system-specific decisions rather than the datasets' ground-truth annotations to define the set of images eligible for attack evaluation. Consequently, only images initially classified as unsafe are included in the robustness analysis.

\begin{table}[h!]
    \centering
    \caption{
        Baseline moderation decisions on clean images.
        $^{*}$For UnsafeBench, the reported values include only the three categories considered in our evaluation: \textit{sexual}, \textit{violence}, and \textit{self-harm}.
    }
    \label{tab:baseline_evaluation}
    \adjustbox{width=\columnwidth}{
        \begin{tabular}{llrrr}
            \toprule
            \textbf{Dataset}
            & \textbf{Moderation service}
            & \textbf{Images}
            & \textbf{Initially safe}
            & \textbf{Initially unsafe} \\
            \midrule
            \multirow{3}{*}{LSPD}
                & OpenAI \texttt{omni-moderation}
                & \num{1000}
                & \num{88} (\num{8.8}\%)
                & \num{912} (\num{91.2}\%) \\
                & Amazon Rekognition
                & \num{1000}
                & \num{38} (\num{3.8}\%)
                & \num{962} (\num{96.2}\%) \\
                & Google Cloud SafeSearch
                & \num{1000}
                & \num{15} (\num{1.5}\%)
                & \num{985} (\num{98.5}\%) \\
            \midrule
            UnsafeBench$^{*}$
                & OpenAI \texttt{omni-moderation}
                & \num{3010}
                & \num{2306} (\num{76.6}\%)
                & \num{704} (\num{23.4}\%) \\
            \midrule
            Hateful Memes
                & OpenAI \texttt{omni-moderation}
                & \num{11605}
                & \num{10268} (\num{88.5}\%)
                & \num{1337} (\num{11.5}\%) \\
            \bottomrule
        \end{tabular}
    }
\end{table}

\subsection{RQ3 Successful and Unsuccessful Cases}
\label{app:rq3_coverage}

For RQ3, we separate images for which at least one tested intensity causes an \texttt{unsafe}-to-\texttt{safe} transition from those that remain \texttt{unsafe} across the entire tested intensity range. Since a minimum successful intensity is undefined for the latter, only images in the first group are included in the minimum-successful-intensity distribution reported in the main paper. Table~\ref{tab:rq3_bypass_coverage} reports, for each intensity-dependent transformation, the proportion of images in the two groups.

\begin{table}[h!]
    \centering
    \caption{
        Proportions of LSPD images that undergo at least one successful \texttt{unsafe}-to-\texttt{safe} transition or remain \texttt{unsafe} across all tested intensities.
    }
    \label{tab:rq3_bypass_coverage}
    \small
    \begin{adjustbox}{width=0.75\columnwidth}
        \begin{tabular}{lrr}
            \toprule
            \textbf{Transformation}
            & \textbf{Changed at least once}
            & \textbf{Never changed} \\
            \midrule
            Box Blur
                & \num{85.96}\%
                & \num{14.04}\% \\
            Gaussian Blur
                & \num{89.14}\%
                & \num{10.86}\% \\
            Gaussian Noise
                & \num{100.00}\%
                & \num{0.00}\% \\
            Salt and Pepper
                & \num{99.56}\%
                & \num{0.44}\% \\
            \bottomrule
        \end{tabular}
    \end{adjustbox}
\end{table}

\subsection{Additional Qualitative Examples}
\label{app:qualitative_examples}

\begin{figure}[ht!]
    \centering
    \begin{subfigure}{0.49\columnwidth}
        \centering
        \includegraphics[width=\linewidth]{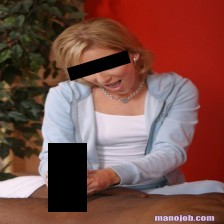}
        \caption{Original image}
    \end{subfigure}
    \hfill
    \begin{subfigure}{0.49\columnwidth}
        \centering
        \includegraphics[width=\linewidth]{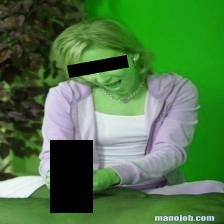}
        \caption{Transformed image}
    \end{subfigure}
    \caption{Successful \texttt{unsafe}-to-\texttt{safe} transition using Split-Merge RGB against Google Cloud SafeSearch.}
    \label{fig:split_merge_example_supp}
\end{figure}

\begin{figure}[ht!]
    \centering
    \begin{subfigure}{0.49\columnwidth}
        \centering
        \includegraphics[width=\linewidth]{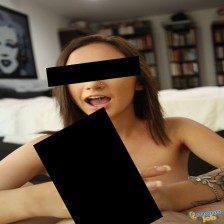}
        \caption{Original image}
    \end{subfigure}
    \hfill
    \begin{subfigure}{0.49\columnwidth}
        \centering
        \includegraphics[width=\linewidth]{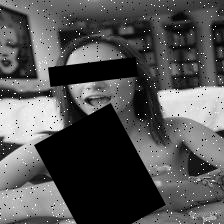}
        \caption{Transformed image}
    \end{subfigure}
    \caption{Successful \texttt{unsafe}-to-\texttt{safe} transition using Salt-and-Pepper Noise against OpenAI \texttt{omni-moderation}.}
    \label{fig:salt_pepper_example_2_supp}
\end{figure}

\begin{figure}[ht!]
    \centering
    \begin{subfigure}{0.49\columnwidth}
        \centering
        \includegraphics[width=\linewidth]{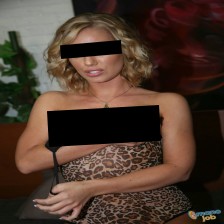}
        \caption{Original image}
    \end{subfigure}
    \hfill
    \begin{subfigure}{0.49\columnwidth}
        \centering
        \includegraphics[width=\linewidth]{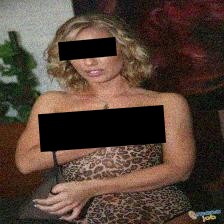}
        \caption{Transformed image}
    \end{subfigure}
    \caption{Successful \texttt{unsafe}-to-\texttt{safe} transition using Gaussian Noise against OpenAI \texttt{omni-moderation}.}
    \label{fig:gaussian_noise_example_supp}
\end{figure}

\begin{figure}[ht!]
    \centering
    \begin{subfigure}{0.49\columnwidth}
        \centering
        \includegraphics[width=\linewidth]{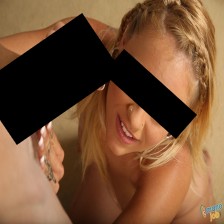}
        \caption{Original image}
    \end{subfigure}
    \hfill
    \begin{subfigure}{0.49\columnwidth}
        \centering
        \includegraphics[width=\linewidth]{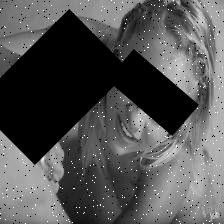}
        \caption{Transformed image}
    \end{subfigure}
    \caption{Successful \texttt{unsafe}-to-\texttt{safe} transition using Salt-and-Pepper Noise against OpenAI \texttt{omni-moderation}.}
    \label{fig:salt_pepper_example_7_supp}
\end{figure}

\begin{figure}[ht!]
    \centering
    \begin{subfigure}{0.49\columnwidth}
        \centering
        \includegraphics[width=\linewidth]{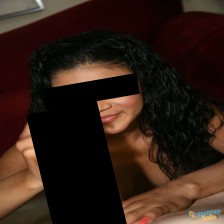}
        \caption{Original image}
    \end{subfigure}
    \hfill
    \begin{subfigure}{0.49\columnwidth}
        \centering
        \includegraphics[width=\linewidth]{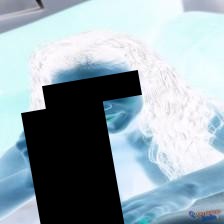}
        \caption{Transformed image}
    \end{subfigure}
    \caption{Successful \texttt{unsafe}-to-\texttt{safe} transition using Color Inversion against OpenAI \texttt{omni-moderation}.}
    \label{fig:invert_color_example_supp}
\end{figure}

\begin{figure}[ht!]
    \centering
    \begin{subfigure}{0.49\columnwidth}
        \centering
        \includegraphics[width=\linewidth]{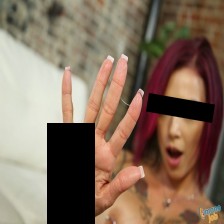}
        \caption{Original image}
    \end{subfigure}
    \hfill
    \begin{subfigure}{0.49\columnwidth}
        \centering
        \includegraphics[width=\linewidth]{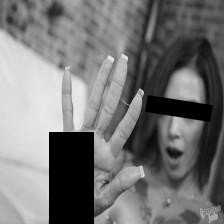}
        \caption{Transformed image}
    \end{subfigure}
    \caption{Successful \texttt{unsafe}-to-\texttt{safe} transition using Grayscale Conversion against Amazon Rekognition.}
    \label{fig:grayscale_example_supp}
\end{figure}

Figures~\ref{fig:split_merge_example_supp}--\ref{fig:grayscale_example_supp}
show additional successful \texttt{unsafe}-to-\texttt{safe} transitions across the evaluated transformations and moderation services. All examples are censored for presentation purposes only; the uncensored images were used for the moderation evaluations.

\end{document}